\documentclass{article}
\usepackage{ijcai26}

\usepackage{times}
\usepackage{soul}
\usepackage{url}
\usepackage[hidelinks]{hyperref}
\usepackage[utf8]{inputenc}
\usepackage[small]{caption}
\usepackage{graphicx}
\usepackage{amsmath}
\usepackage{amsthm}
\usepackage{booktabs}
\usepackage{algorithm}
\usepackage{algorithmic}
\usepackage[switch]{lineno}

\usepackage{subcaption}  

\usepackage{forest}
\usepackage{colortbl} 
\usepackage{multirow} 
\definecolor{Gray}{gray}{0.94}
\usepackage{graphicx} 
\usepackage{subcaption}  
\NewDocumentCommand\emojidizzy{}{
        \includegraphics[scale=0.0245]{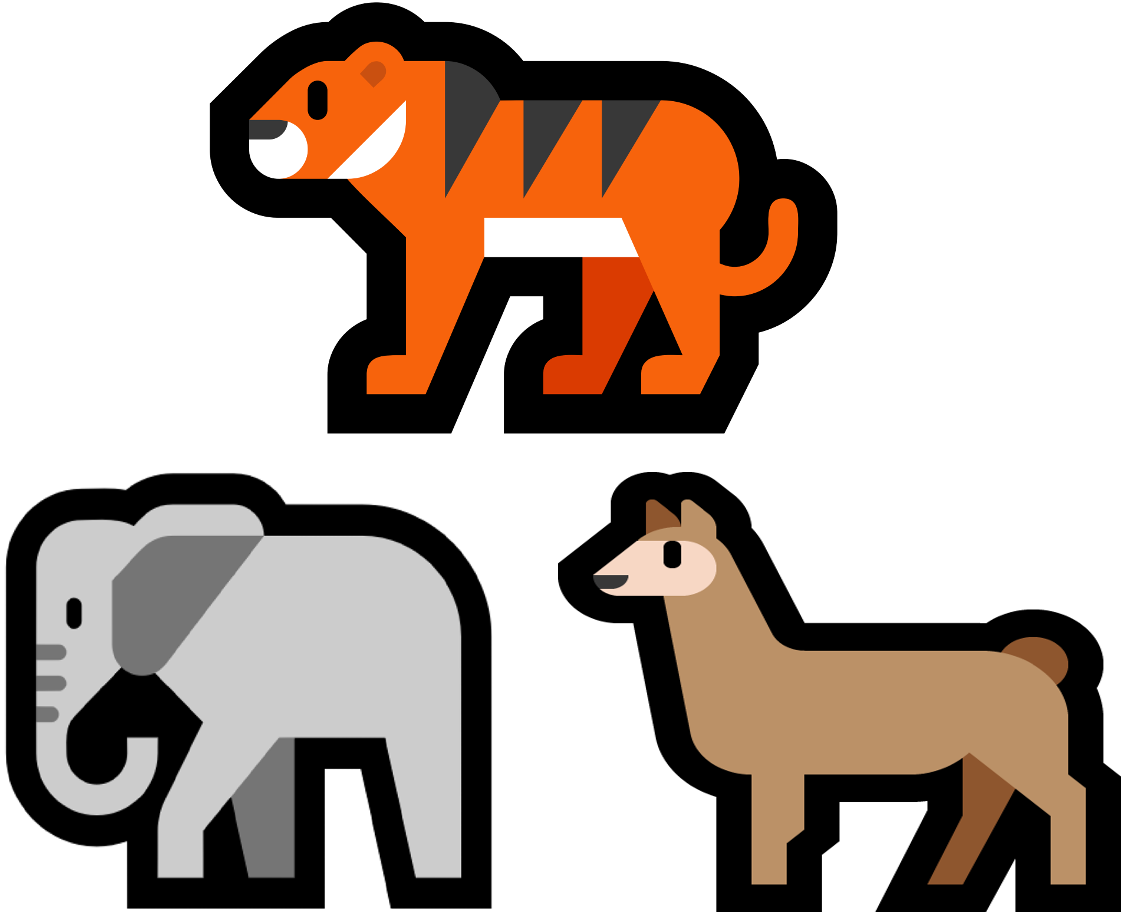}
}

\usepackage{amsmath}

\usepackage{pifont}
\newcommand{\cmark}{\ding{51}}%
\newcommand{\xmark}{\ding{55}}%
\usepackage{threeparttable} 
\usepackage{fontawesome}
\usepackage{booktabs}
\definecolor{harvestgold}{RGB}{173, 185, 202}
\definecolor{DarkGreen}{RGB}{196, 224, 180}
\usepackage{tikz}

\definecolor{academicblue}{RGB}{114, 154, 213}
\hypersetup{
    colorlinks=true,       
    linkcolor=academicblue, 
    citecolor=academicblue, 
}
\PassOptionsToPackage{html}{xcolor}
\usepackage{xcolor} 
\usepackage{tikz}   
\usepackage[most]{tcolorbox} 
\usepackage{cleveref}
\usepackage{enumitem}
\definecolor{harvestgold}{RGB}{173, 185, 202}
\definecolor{DarkGreen}{RGB}{196, 224, 180}
\definecolor{academicblue}{HTML}{000080}
\definecolor{selfevolagent_dark}{HTML}{37D2A6}
\definecolor{selfevolagent_light}{HTML}{9BE9D3}
\definecolor{selfevolagent_lighter}{HTML}{CDF4E9}
\definecolor{colorBlue}{HTML}{8EAADC}
\definecolor{colorGrey}{HTML}{ADB9CA}
\definecolor{colorGreen}{HTML}{C4E0B4}
\definecolor{colorLavender}{HTML}{E2E5F9} 

\hypersetup{
    colorlinks=true,
    linkcolor=academicblue,
    citecolor=academicblue,
    urlcolor=academicblue
}
\definecolor{subheadgray}{gray}{0.35} 
\definecolor{tableblue}{rgb}{0.867, 0.922, 0.969}
\newcommand{\gc}{\cellcolor{tableblue}}
\usepackage{fix-cm}

\usepackage[most]{tcolorbox}  
\usepackage[framemethod=TikZ]{mdframed} 

\mdfdefinestyle{mystyle}{%
  rightline=true,
  innerleftmargin=10,
  innerrightmargin=10,
  outerlinewidth=3pt,
  topline=false,
  rightline=true,
  bottomline=false,
  skipabove=\topsep,
  skipbelow=\topsep
}
\newtcolorbox{myboxi}[1][]{
    breakable,
    title=#1,
    colback=red!5,
    colbacktitle=red!5,
    coltitle=black,
    fonttitle=\bfseries,
    bottomrule=0pt,
    toprule=0pt,
    leftrule=2pt,
    rightrule=2pt,
    titlerule=0pt,
    arc=0pt,
    outer arc=0pt,
    colframe=red,
    left=0.02mm,
    right=1mm,
    top=1.1mm,
    bottom=0.1mm,
    boxsep=0.5pt,
    toptitle=1mm,
    bottomtitle=1mm,
}

\title{\emojidizzy Harnessing Multiple Large Language Models: A Survey on LLM Ensemble}

\author{
\fontsize{10pt}{10pt}\selectfont 
Zhijun Chen\textsuperscript{\rm 1}, 
Xiaodong Lu\textsuperscript{\rm 1}, 
Jingzheng Li\textsuperscript{\rm 2}, 
Pengpeng Chen\textsuperscript{\rm 3}, 
Zhuoran Li\textsuperscript{\rm 1}, 
Kai Sun\textsuperscript{\rm 4}, 
Yuankai Luo\textsuperscript{\rm 5},  \\
Qianren Mao\textsuperscript{\rm 2}, 
Ming Li\textsuperscript{\rm 6}, 
Likang Xiao\textsuperscript{\rm 1}, 
Dingqi Yang\textsuperscript{\rm 7}, 
Xiao Huang\textsuperscript{\rm 8}, 
Yikun Ban\textsuperscript{\rm 1,*}, 
Hailong Sun\textsuperscript{\rm 1,*}, 
Philip S. Yu\textsuperscript{\rm 9,*}
    \affiliations
    \fontsize{11pt}{10pt}\selectfont 
    \textsuperscript{\rm 1}State Key Laboratory of Complex \&  Critical Software Environment, Beihang University, Beijing, China\\
    \textsuperscript{\rm 2}Zhongguancun Laboratory, Beijing, China\\
    \textsuperscript{\rm 3}Aviation System Engineering Institute of China, Beijing, China\\
    \textsuperscript{\rm 4}Xi’an Jiaotong University, Xi’an, China\\
      \textsuperscript{\rm 5}Nanjing University, Nanjing, China\\
        \textsuperscript{\rm 6}Tsinghua University, Beijing, China\\
    \textsuperscript{\rm 7}University of Macau, Macau SAR, China\\
    \textsuperscript{\rm 8}The Hong Kong Polytechnic University, Hong Kong, China\\
    \textsuperscript{\rm 9}University of Illinois at Chicago, Chicago, USA\\
    \emails{
     \fontsize{10pt}{10pt}\selectfont
    \{zhijunchen, xiaodonglu, jingzhengli, yikunb, sunhl\}@buaa.edu.cn,  
    maoqr@zgclab.edu.cn, 
    psyu@uic.edu \\
    }  
}

\begin{document}

\maketitle

\begin{NoHyper}
\def\thefootnote{}\footnotetext{*Corresponding authors.}
\end{NoHyper}

\begin{abstract}
LLM Ensemble---which involves the comprehensive use of multiple large language models (LLMs), each aimed at handling user queries during downstream inference, to benefit from their individual strengths---has gained substantial attention recently. The widespread availability of LLMs, coupled with their varying strengths and out-of-the-box usability, has profoundly advanced the field of LLM Ensemble. This paper presents the first systematic review of recent developments in LLM Ensemble. First, we introduce our taxonomy of LLM Ensemble and discuss several related research problems. Then, we provide a more in-depth classification of the methods under the broad categories of ``ensemble-before-inference, ensemble-during-inference, ensemble-after-inference'', and review relevant methods. Finally, we introduce related benchmarks and applications, summarize existing studies, and suggest  future research directions. 

\faGithub~\href{https://github.com/junchenzhi/Awesome-LLM-Ensemble}{
  \textcolor[rgb]{0.0, 0.4039, 0.5843}{https://github.com/junchenzhi/Awesome-LLM-Ensemble}
}

\end{abstract}

\section{Introduction}
\label{Introduction}

In recent years, the landscape of artificial intelligence has been dramatically reshaped by the development of Large Language Models (LLMs), including Gemini~\cite{team2023gemini}, GPT-4~\cite{achiam2023gpt}, Llama~\cite{touvron2023llama}, and the recently introduced DeepSeek~\cite{liu2024deepseek}.
The success of these LLMs continues to fuel widespread research enthusiasm, with a remarkable total of over 182,000 large language models now accessible in the Hugging Face library.~\footnote{\url{https://huggingface.co/models}.}

Behind this research enthusiasm, however, we can identify two main aspects: 
1) \textit{The performance concerns}: 
The direct out-of-the-box capability of LLMs (from zero-shot inference) and their indirect out-of-the-box capability (from in-context-learning few-shot inference) still raise performance worries, including accuracy, hallucinations, and misalignment with human intent, among others;
2) \textit{The varying strengths and weaknesses of LLMs, each with different inference costs}: 
Due to differences in architecture,  size, tokenization, dictionary, training data,  methodology, these LLMs exhibit substantial variability and their responses can differ significantly.
With the above two aspects in mind and drawing on the spirit of Ensemble Learning~\cite{dong2020survey,zhou2021ensemble,mohammed2023comprehensive}, it is natural to consider that, for each  query, rather than  relying on a single LLM based on public rankings or other criteria, it might be more advantageous to simultaneously consider multiple LLM candidates (usable out-of-the-box) and harness their distinct strengths.
This is exactly what the  emerging field of \textit{LLM Ensemble} explores.

\begin{figure*}[t!]
    \centering
    \begin{minipage}{0.31\textwidth}
        \centering
        \includegraphics[width=\textwidth]{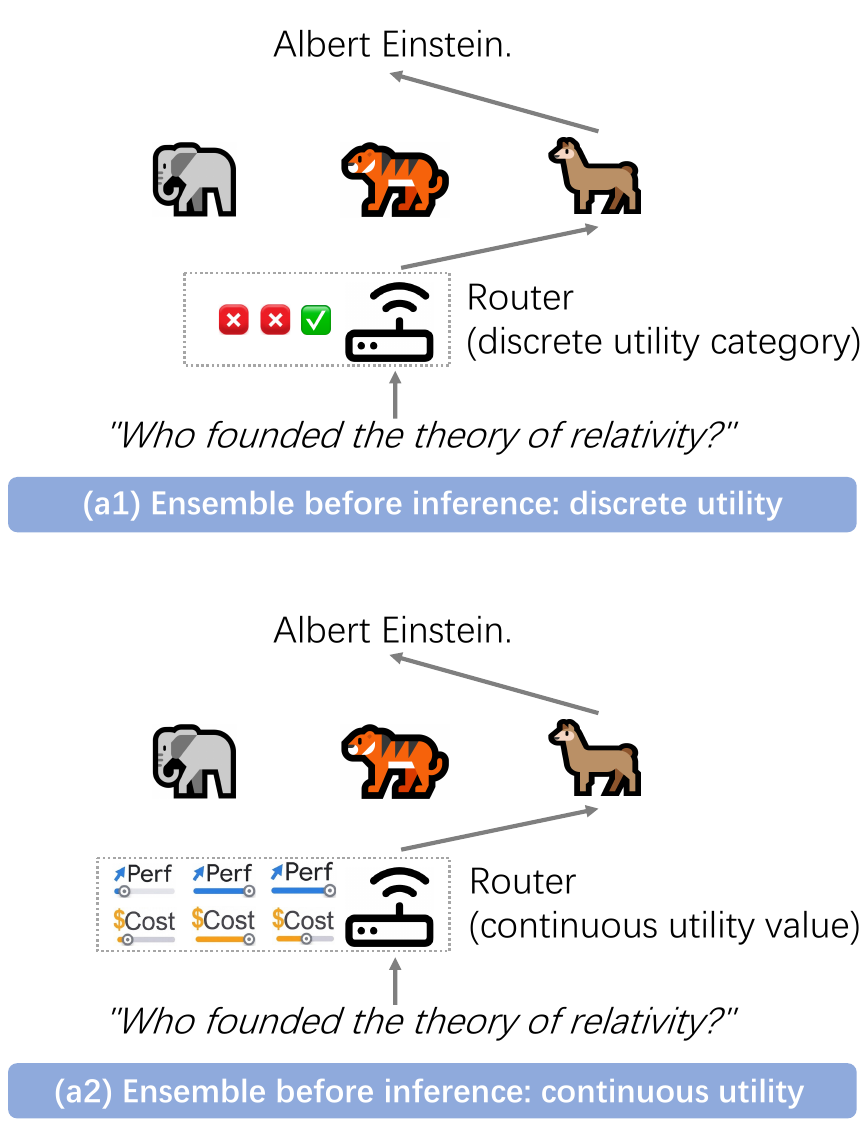}  
        \subcaption{Ensemble before inference.}  
        \label{figure-a}
    \end{minipage}
    \hfill
    \begin{minipage}{0.31\textwidth}
        \centering
        \includegraphics[width=\textwidth]{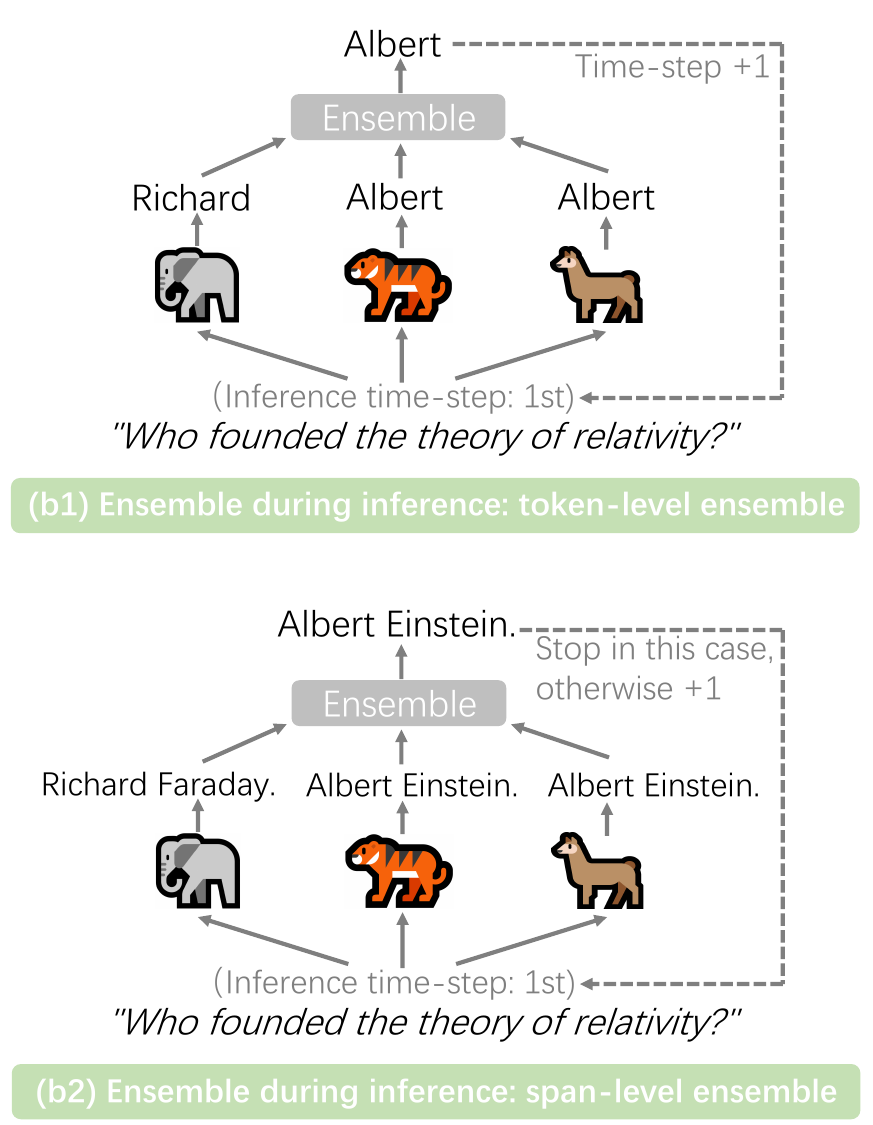}  
        \subcaption{Ensemble during inference.} 
         \label{figure-b}
    \end{minipage}
    \hfill
    \begin{minipage}{0.31\textwidth}
        \centering
        \includegraphics[width=\textwidth]{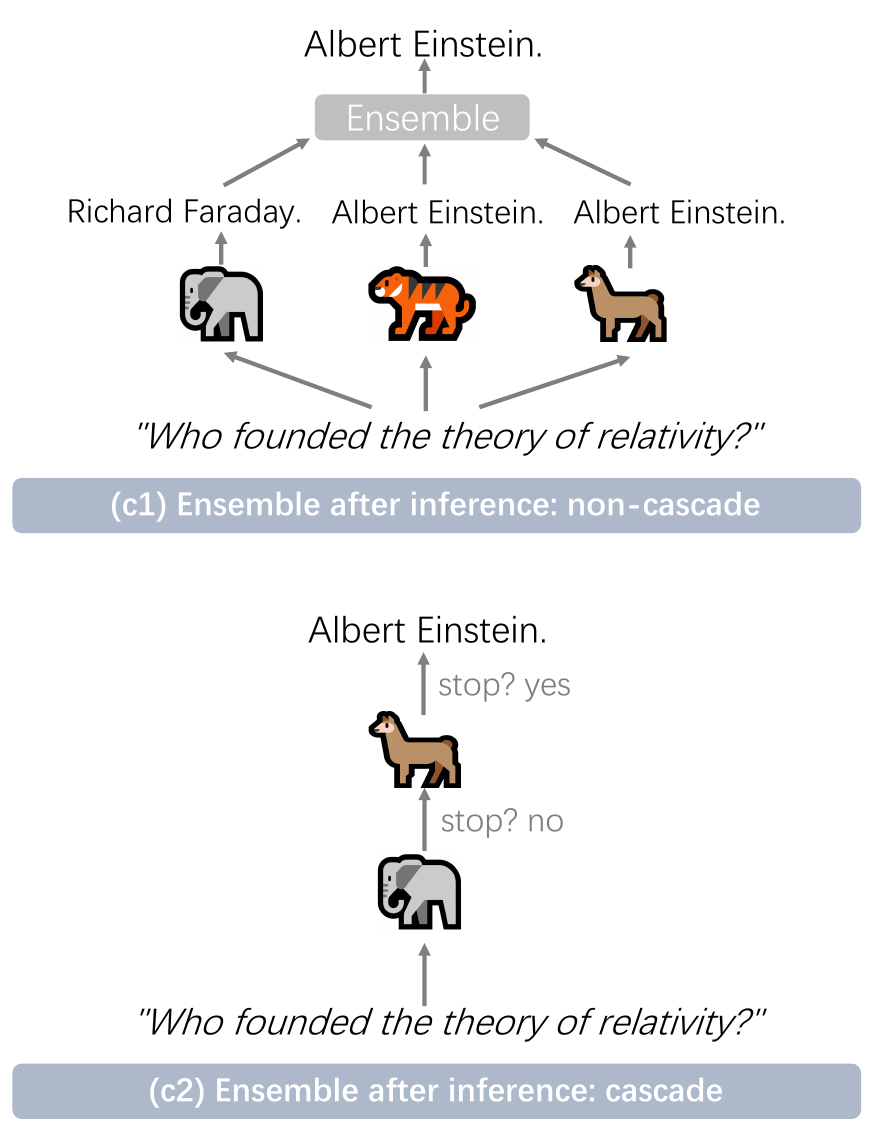}  
        \subcaption{Ensemble after inference.}  
         \label{figure-c}
    \end{minipage}
    \caption{Illustration of the LLM Ensemble taxonomy. (Note that for \textit{(b) ensemble-during-inference}, there is also a \textit{(b3) process-level ensemble} approach not depicted in the figure, 
     due to layout considerations and the status that this approach 
     is rarely instantiated.)}
    \label{figure-1}
\end{figure*}

Existing LLM Ensemble methods can be broadly categorized into three types, depending on the sequence of \textit{LLM inference} and \textit{ensemble}:
1) 
\textit{Ensemble-before-inference} approach, utilizes the given query information while considering the diverse characteristics of all LLM candidates to route an appropriate model for inference (this approach is similar to the \textit{hard voting}  strategy in Ensemble Learning);
2)
\textit{Ensemble-during-inference} approach, aggregates incomplete responses (e.g., token-level information) from multiple LLMs during the decoding process and feeds the combined result back into all the models;
3)
\textit{Ensemble-after-inference} approach, performs the ensemble after full responses (instead of fragments) have been generated by all models or a subset of them.
Despite the emergence of numerous methods derived from these paradigms recently, there is still no formal survey that offers a comprehensive review of the core ideas and related research.



We present the first comprehensive survey on LLM Ensemble, introducing recent advances and focusing on taxonomy, related problems, methods, benchmarks, applications, and future directions. 
We hope that this survey will provide a thorough review for researchers and inspire further exploration.

\section{LLM Ensemble Taxonomy and Related Problems}

\subsection{LLM Ensemble Taxonomy}
\label{Taxonomy-section}

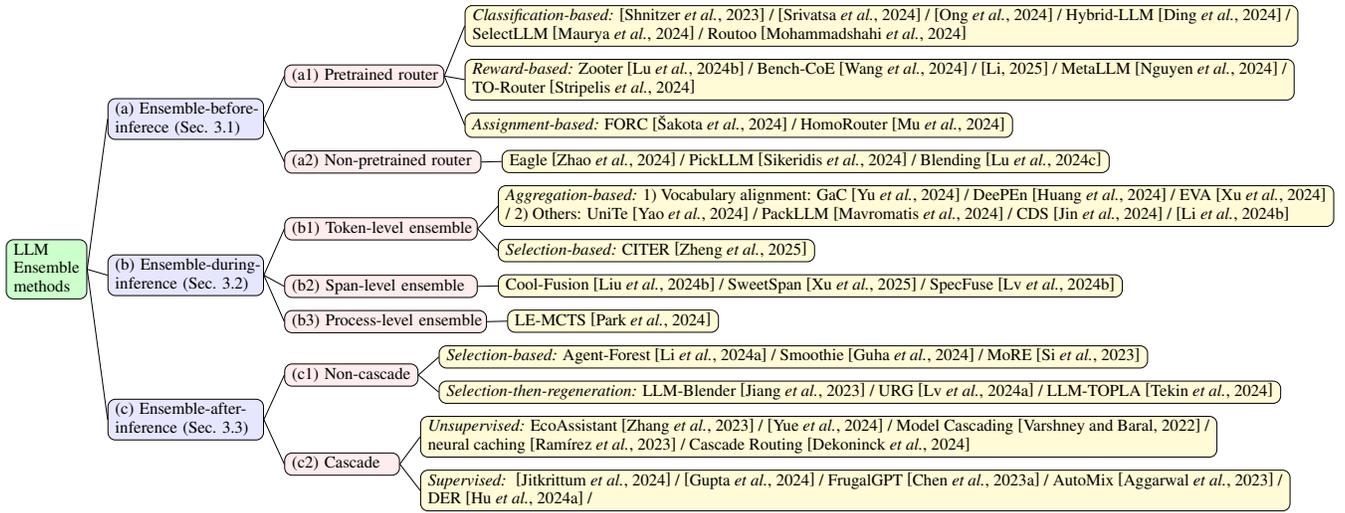
\begin{figure*}[t!]
	\centering
	\resizebox{\textwidth}{!}{
	\begin{forest}
  for tree={
  grow=east,
  reversed=true,
  anchor=base west,
  parent anchor=east,
  child anchor=west,
  base=left,
  font=\small,
  rectangle,
  draw,
  rounded corners,align=left,
  minimum width=2.5em,
  inner xsep=4pt,
  inner ysep=1pt,
  },
  where level=1{text width=5em,fill=blue!10}{},
  where level=2{text width=5em,font=\footnotesize,fill=pink!30}{},
  where level=3{font=\footnotesize,yshift=0.26pt,fill=yellow!20}{},
  [LLM \\ Ensemble\\ methods,fill=green!20
        [(a) Ensemble-before-\\inference (Sec.~\ref{Ensemble-before-inference}), text width=7.6em
          [(a1) Discrete utility, text width=8.em
            [\textit{Classification-based methods}: 
              \cite{shnitzer2023large} / \cite{srivatsa2024harnessing} / SelectLLM~\cite{maurya2024selectllm} / Routoo~\cite{mohammadshahi2024routoo}\\
              CSCR~\cite{shirkavand2025cost} / RADAR~\cite{fernandez2025radar} / FORC~\cite{vsakota2024fly}
            ]
            [\textit{Comparison-based methods}:
              \cite{ong2025routellm} / Hybrid-LLM~\cite{dinghybrid} / P2L~\cite{frick2025prompt} / \cite{zhang2025leveraging}\\
              Meta-Router~\cite{zhang2025meta} / Eagle~\cite{zhao2024eagle} / Bench-CoE~\cite{wang2024bench}
            ]
          ]
          [(a2) Continuous utility, text width=8.2em
            [\cite{li2025llm} / MetaLLM~\cite{nguyen2024metallm} / MixLLM~\cite{wang2025mixllm} / IRT-Router~\cite{song2025irt} /
             Avengers~\cite{zhang2025avengers} /
             \cite{li2025rethinking} \\ OmniRouter~\cite{mei2025omnirouter}  / \cite{wu2025efficient} /
             TO-Router~\cite{stripelis2024tensoropera} / HomoRouter~\cite{mu2024adaptive} / PickLLM~\cite{sikeridis2024pickllm}
            ]
          ]
        ]
        [(b) Ensemble-during-\\inference (Sec.~\ref{Ensemble-during-inference}),text width=7.6em
            [(b1) Token-level ensemble,text width=9.6em
              [\emph{Aggregation-based normal methods:} 1) Vocabulary alignment: GaC~\cite{yu2024breaking} / DeePEn~\cite{huang2024ensemble} / EVA~\cite{xu2024bridging}  \\ / 2) Others: UniTe~\cite{yao2024determine} / PackLLM~\cite{mavromatis2024pack} 
              ]
              [\emph{Aggregation-based specific-goal methods:} \cite{li2024purifying} / DeRa~\cite{liu2024decoding}  / MOD~\cite{shi2024decoding} 
              ]
              [\emph{Aggregation-based fine-tuning methods:}  Copilot~\cite{zou2025transformer} / 
              LLMBoost~\cite{chen2025llmboost} / 
              UltraFuser
              ~\cite{ding2024mastering} 
              ]
              [\emph{Selection-based methods:} 
              CDS~\cite{jin2024collaborative} /
              Co-Llm~\cite{shen2024learning} / 
               CITER~\cite{zheng2025citer} /  
               ABE~\cite{wicks2025token}
              ]
          ]
          [(b2) Span-level  ensemble, text width=9.6em
            [\emph{Assessment-based methods:} Cool-Fusion~\cite{liu2025cool} / SweetSpan~\cite{xu2025hit} / SpecFuse~\cite{lv2024specfuse}
            ]
                   [\emph{Others:} 
                        CoS~\cite{fu2025fast} 
                  ]
          ]
            [(b3) Process-level ensemble, text width=10.1em
            [ LE-MCTS~\cite{park2025ensembling} 
            ]
          ]
        ]
        [(c) Ensemble-after-\\inference (Sec.~\ref{Ensemble-after-inference}),text width=7.6em
          [(c1) Non-cascade,text width=6.4em
            [\emph{Selection-based:} Agent-Forest~\cite{li2024more} / Smoothie~\cite{guhasmoothie}  /  LLM-PeerReview~\cite{chen2025scoring} / MoRE~\cite{si2023getting} ]
             [\emph{Selection-then-regeneration:} LLM-Blender~\cite{jiang2023llm}  / LLM-TOPLA~\cite{tekin2024llm} / URG~\cite{lv2024urg} ]
            ]
          [(c2) Cascade, text width=5.4em
            [\emph{Unsupervised:} EcoAssistant~\cite{zhang2023ecoassistant} / \cite{yuelarge} / Model Cascading~\cite{varshney2022model}   / \\  neural caching~\cite{ramirez2023cache} / Cascade Routing~\cite{dekoninck2024unified} ] 
             [\emph{Supervised:}   ~\cite{jitkrittum2024does} / FrugalGPT~\cite{chenfrugalgpt} / \cite{gupta2024language} / AutoMix~\cite{aggarwal2023automix} /   DER~\cite{hu2024dynamic}    ]
            ]
        ]
    ]
\end{forest}
	}
	\caption{Taxonomy of LLM Ensemble methods.}
	\label{Fig: Taxonomy}
\end{figure*}

This section formally introduces our LLM Ensemble taxonomy, illustrated by the schematic shown in Figure~\ref{figure-1} and the detailed taxonomy in Figure~\ref{Fig: Taxonomy}.
As mentioned in Section~\ref{Introduction}, the following three broad categories of LLM Ensemble exist.

\textbf{(a) Ensemble before inference.}
In essence, this approach employs a routing algorithm prior to LLM inference to allocate a specific query to the most suitable model, allowing the selected model that is specialized for the query and typically more cost-efficient inference to perform the task.
As illustrated in Figure~\ref{figure-a} and Figure~\ref{Fig: Taxonomy}, existing methods can be classified into two categories, depending on whether the router utility is discretized: 
(a1) \textit{discrete utility methods} and (a2) \textit{continuous utility methods}.

\textbf{(b) Ensemble during inference.}
As the most granular form of ensemble among the three broad categories, this type of approach encompasses: (b1) \textit{token-level ensemble} methods, integrate the token-level outputs of multiple models at the finest granularity of decoding;
(b2) \textit{span-level ensemble methods}, perform ensemble at the level of a sequence fragment (e.g., a span of four words); 
(b3) 
\textit{process-level ensemble methods}, select the optimal reasoning process step-by-step within the reasoning chain for a given complex reasoning task. 
\textit{Note that for these ensemble-during-inference methods, the aggregated text segments will be concatenated with the preceding text and fed back into the models.}

\textbf{(c) Ensemble after inference.}
These methods can be classified into two categories: 
(c1) \textit{Non-cascade methods}, perform ensemble by integrating multiple complete responses contributed from all LLM candidates;
(c2) \textit{Cascade methods}, consider both performance and inference costs, progressively performing inference through a chain of LLM candidates ranked primarily by model size to identify the most suitable response and terminate the cascade process.

\subsection{Related Problems}
Here we briefly introduce the closely related problems.

\textbf{LLM Merging}, also known as \textbf{LLM Fusion}~\cite{yang2024model,yang2025cabs,ruan2025task,li2023deep}, integrates parameters from various LLMs to construct a universal model without requiring original training data or extensive computation. 
It is highly relevant to LLM Ensemble, as both promote knowledge fusion and transfer.

\textbf{LLM  Collaboration} leverages the distinct strengths of each model to approach tasks with increased flexibility~\cite{du2023improving,lu2024merge,chen2025survey,tran2025multi,zhang2025if,li2024survey,guo2024large}. 
Unlike LLM Ensemble where models are employed with equal status to \textit{directly faced to user queries}, the collaboration approach assigns distinct roles to each LLM, exchanging response information to enhance task resolution.
Furthermore, a closely related line of research to LLM Collaboration—which mainly focuses on zero-shot inference—is \textbf{Multi-LLM Reinforcement Learning}~\cite{sun2024llm,jin2025comprehensive,li2026adaptive}, which investigates how multiple agents learn optimal decision-making strategies in a shared environment.

\textbf{Weak Supervision}~\cite{zhang2021wrench,chen2023neural,zhang2022survey,ratner2017snorkel}, 
sometimes referred to as
\textbf{Learning from Crowds}~\cite{chen2021structured,chen2022adversarial,zhang2022knowledge}, uses weak labels contributed from multiple weak annotation sources to perform information aggregation~\cite{chen2023black,zheng2017truth} (corresponding \textit{non-cascade ensemble after inference} in LLM Ensemble) or directly train a classifier~\cite{zhang2021wrench}, yet most methods focus on classification
rather than general open-ended generation tasks.

\section{Methodology}
\label{Methodology}
In this section, following the taxonomy in Section~\ref{Taxonomy-section}, we systematically review the three types of methods—ensemble before inference, ensemble during inference and ensemble after inference—in Sections~\ref{Ensemble-before-inference}, ~\ref{Ensemble-during-inference}, and ~\ref{Ensemble-after-inference}, respectively.

\subsection{Ensemble Before Inference}
\label{Ensemble-before-inference}

Since the ensemble-before-inference methods require routing a query to the most suitable LLM before LLM inference, the core of such methods lies in predicting the utility of candidate models for a given query under certain preferences (e.g., performance or cost). Based on how they formulate the utility of candidate LLMs, we divide existing methods into two categories of \textit{discrete utility methods} and \textit{continuous utility methods}. A comprehensive summary of the ensemble before inference methods is presented in Table~\ref{tab:Ensemble-before-inference}.

\subsubsection{3.1.1 (a1) Discrete Utility Methods}
\label{Classification-Based Methods}

Discrete utility methods discretize the model utility into \textit{categorical labels}. Depending on whether the categories are defined independently or via pairwise comparisons, these methods can be further divided into \textit{classification-based} and 
\textit{comparison-based methods}.

\paragraph{Classification-based methods.} 
For this type of method, the utility of a candidate LLM for a given query is typically treated as a discrete variable reflecting the quality of the model’s response. Typically, the utility is binary: a value of 1 indicates that the model produces a satisfactory response, whereas 0 indicates an unsatisfactory one. For example, in a commonsense question-answering (QA) task, a correct answer is labeled as 1 and an incorrect answer as 0 (e.g., \cite{shnitzer2023large}). Under this formulation, the routing problem naturally becomes a multi-label binary classification task, where the router estimates the probability that each candidate LLM will produce a satisfactory output. Based on different optimization objectives, one can then design corresponding decision strategies—for instance, when model costs are known, the router can compute a combined value via a weighted aggregation of the predicted score and cost \cite{fernandez2025radar}, and select the model with the highest combined value to balance performance and cost. Following this line, existing methods extend the classification-based methods along \textit{empirical analysis} \cite{srivatsa2024harnessing}, \textit{decision policy} \cite{maurya2024selectllm,shnitzer2023large,vsakota2024fly}, and \textit{router architecture} \cite{shirkavand2025cost,mohammadshahi2024routoo}.


\paragraph{Comparison-based methods.} Compared to multi-label binary classification routing, which requires estimating an absolute (pointwise) utility score for each candidate LLM’s output, comparison-based approaches only need to infer relative preferences among a pair of model outputs, thereby simplifying the routing objective and reducing supervision difficulty \cite{zhang2025meta}. Specifically, a preference datum can be represented as a tuple $(q, M_1, M_2, y)$, where $q$ denotes a given query, $M_1$ and $M_2$ are two candidate LLMs, and $y$ is a binary label such that $y=1$ indicates that the response of $M_1$ is better than that of $M_2$, while $y=0$ indicates otherwise (e.g., \cite{ong2025routellm}). In this setting, the router aims to predict the win rate of $M_1$, i.e., the probability that $M_1$’s response is better given the query $q$ and the model pair $(M_1, M_2)$. Following this idea, subsequent methods extend the comparison-based routing framework along several directions, including \textit{data construction} \cite{ong2025routellm,zhang2025leveraging}, \textit{learning target} \cite{frick2025prompt,chen2025tagrouter,wang2024bench}, \textit{routing strategy} \cite{zhang2025meta,zhao2024eagle}

\setlength{\cmidrulewidth}{0.01em}  
\begin{table*}[t!]
\hspace*{1.3cm}  
\scalebox{0.586}{
\fontsize{9pt}{9pt}
\begin{threeparttable}
\begin{tabular}{llcc c c c c}
\toprule
& \textbf{Methods}    &  \textbf{Param./Non-param.$^{\S}$} & \textbf{Goals}   & \textbf{Loss functions} & \textbf{Tasks}$^{*}$  &  \textbf{Generalization$^\dag$}    & \textbf{Code} \\
\midrule
\multirow{ 14 }{*}{(a1)}   &
\gc \cite{shnitzer2023large}  & \gc Parametric & \gc Performance   & \gc Binary CE loss & \gc OE-G/EM-G & \gc \xmark    & \gc -\\
  &
\cite{srivatsa2024harnessing}  & Parametric & Performance   & Class-balanced CE loss & EM-G & \cmark    & \href{https://github.com/kvadityasrivatsa/llm-routing}{\textcolor{gray}{[Link]}}\\
  &
\gc SelectLLM \cite{maurya2024selectllm}  & \gc Parametric & \gc Performance and cost   & \gc Class-balanced CE loss & \gc EM-G & \gc \xmark    & \gc -\\
  &
Routoo~\cite{mohammadshahi2024routoo}  & Parametric & Performance and cost   & CE loss & EM-G & \xmark    & -\\
  &
\gc CSCR~\cite{shirkavand2025cost}  & \gc Parametric & \gc Performance and cost   & \gc InfoNCE & \gc OE-G/EM-G & \gc \xmark    & \gc -\\
  &
RADAR~\cite{fernandez2025radar}  & Parametric & Perf., cost and query time   & Binary CE loss &  EM-G &  \xmark    &  -\\
 &
\gc FORC~\cite{vsakota2024fly}  & \gc Parametric & \gc Performance and cost   & \gc Avg. of multi-metric integration & \gc OE-G/EM-G & \gc \cmark    & \gc \href{https://github.com/epfl-dlab/forc}{\textcolor{gray}{[Link]}}\\
  &
\cite{ong2025routellm}  & Parametric & Performance and cost   & Binary CE loss & OE-G/EM-G & \cmark    & \href{https://github.com/lm-sys/RouteLLM}{\textcolor{gray}{[Link]}}\\
  &
\gc Hybrid-LLM~\cite{dinghybrid}  & \gc Parametric & \gc Performance and cost   & \gc Binary CE loss & \gc OE-G/EM-G & \gc \xmark    & \gc \href{https://github.com/m365-core/hybrid_llm_routing}{\textcolor{gray}{[Link]}}\\
  &
P2L~\cite{frick2025prompt}  & Parametric & Performance and cost   & Binary CE loss & OE-G/EM-G & \xmark    & \href{https://github.com/lmarena/p2l}{\textcolor{gray}{[Link]}}\\
  &
\gc \cite{zhang2025leveraging}  & \gc Parametric & \gc Performance and cost   & \gc Binary CE loss & \gc OE-G/EM-G & \gc \xmark    & \gc -\\
  &
 Meta-Router~\cite{zhang2025meta}  & Non-parametric &  Performance and cost   & - &  OE-G/EM-G &  \xmark    &  \\
  &
\gc Eagle~\cite{zhao2024eagle}  & \gc Non-parametric & \gc Performance and cost   & \gc - & \gc OE-G/EM-G & \gc \xmark    & \gc -\\
  &
Bench-CoE~\cite{wang2024bench}  &  Parametric & Performance   & CE loss &  EM-G &  \cmark    & \href{https://github.com/ZhangXJ199/Bench-CoE}{\textcolor{gray}{[Link]}} \\
\cmidrule{2-8}
\multirow{ 11 }{*}{(a2)}   &
\gc \cite{li2025llm}  & \gc Parametric & \gc Performance and cost   & \gc MSE loss & \gc OE-G/EM-G & \gc \cmark    & \gc -\\
  &
MetaLLM~\cite{nguyen2024metallm}  & Parametric & Performance   & MSE loss & EM-G & \xmark    & \href{https://github.com/mail-research/MetaLLM-wrapper/}{\textcolor{gray}{[Link]}}\\
  &
\gc MixLLM~\cite{wang2025mixllm}  & \gc Parametric & \gc Perf., cost and query time   & \gc Negative Expected Reward
 & \gc OE-G/EM-G & \gc \xmark    & \gc -\\
  &
IRT-Router~\cite{song2025irt}  & Parametric & Perf., cost and query time   & Binary CE loss & OE-G/EM-G & \xmark    & \href{https://github.com/Mercidaiha/IRT-Router}{\textcolor{gray}{[Link]}}\\
  &
\gc Avengers~\cite{zhang2025avengers}  & \gc Non-parametric & \gc Performance   & \gc Clustering & \gc OE-G/EM-G & \gc \xmark    & \gc \href{https://github.com/ZhangYiqun018/Avengers}{\textcolor{gray}{[Link]}}\\
  &
OmniRouter~\cite{mei2025omnirouter}  & Parametric & Performance and cost   & MSE loss, Binary CE loss & EM-G & \xmark    & \href{https://github.com/agiresearch/OmniRouter}{\textcolor{gray}{[Link]}}\\
  &
\gc \cite{li2025rethinking}  & \gc Non-parametric & \gc Performance and cost   & \gc - & \gc OE-G/EM-G & \gc \xmark    & \gc -\\
  &
\cite{wu2025efficient}  & Non-parametric & Performance and cost   & - &  OE-G/EM-G & \xmark    & \href{https://github.com/fzwark/PORT}{\textcolor{gray}{[Link]}}\\
  &
\gc TO-Router~\cite{stripelis2024tensoropera}  & \gc Parametric & \gc Perf., cost and query time   & \gc KL divergence & \gc \gc OE-G/EM-G & \gc \cmark    & \gc -\\
  &
 HomoRouter~\cite{mu2024adaptive}  & Parametric &  Performance and cost   & MSE loss &  EM-G &  \cmark    & -\\
  &
\gc PickLLM~\cite{sikeridis2024pickllm}  & \gc Parametric & \gc Perf., cost and query time   & \gc Negative Expected Reward
 & \gc OE-G/EM-G & \gc \cmark    & \gc -\\

\bottomrule
\end{tabular}

\begin{tablenotes}
\small
\item[1] $^{\S}$: ``Param./Non-param.$^{\S}$'' indicates whether the router have learnable parameters.
\item[2] $^{*}$: OE-G denotes Open-Ended Generation; EM-G denotes Exact-Match Generation, characterized by objectively verifiable answers (e.g., mathematical solutions).
\item[3] $\dag$:  It denotes whether the router can generalize to new domains.
\end{tablenotes}

\end{threeparttable}
}
\captionsetup{type=table,skip=2pt}

\caption{Summary of ensemble-before-inference methods.
}
\label{tab:Ensemble-before-inference}
\end{table*}

\subsubsection{3.1.2 (a2) Continuous Utility Methods}
\label{Reward-Based Methods}
Unlike prior discrete-category approaches, continuous-utility methods model LLM utility as real-valued variables, such as response length or performance scores. This formulation enables a fine-grained characterization of model behavior, capturing subtle performance differences obscured by categorical definitions. Moreover, continuous utilities naturally aggregate multiple objectives (e.g., latency and cost) into a unified scalar, allowing the router to jointly optimize these objectives and learn more flexible policies. The most straightforward continuous-utility approach formulates routing as a utility prediction problem, where the router is often trained to predict the utility of each candidate model using regression-based losses \cite{song2025irt,zhang2025avengers,mei2025omnirouter,li2025rethinking,wu2025efficient,mu2024adaptive}. For example, OmniRouter~\cite{mei2025omnirouter} trains two MLP predictors to estimate performance and cost, respectively, and then considers batch query routing as a cost-constrained performance optimization problem, thereby achieving a trade-off between performance and cost. Beyond such explicit utility modeling, other methods instead directly learn a routing policy that outputs selection probabilities over candidate LLMs, treating utility as a feedback signal to optimize the policy \cite{nguyen2024metallm,li2025llm,lu2024routing,stripelis2024tensoropera,sikeridis2024pickllm}. For example, Li et al.~\shortcite{li2025llm} define the reward as a weighted sum of cost and performance, and adopt a multi-objective policy gradient algorithm to learn the selection policy.

\setlength{\cmidrulewidth}{0.01em}  
\begin{table*}[t]
\hspace*{0.4cm}  
\scalebox{0.61}{
\fontsize{9pt}{9pt}
\begin{threeparttable}  
\scalebox{0.95}{
\begin{tabular}{llcc c c c c c c }
\toprule
 &\textbf{Methods}    &  \textbf{Granularities} &\textbf{Main Strategies}   &\textbf{Unsup.}  & \textbf{\# Models}$^\dag$  & \textbf{Other Traits$^{\S}$}  & \textbf{Efficiency-Aware}  & \textbf{Tasks}$^{*}$ & \textbf{Code} \\

\midrule
\multirow{15}{*}{(b1)} &
\gc GaC~\cite{yu2024breaking} &\gc Token  & \gc Averaging agg.,Vocabulary align.  &\gc \cmark   &\gc $\geq 2$   &\gc Union dictionary  &\gc \xmark  & \gc OE-G/EM-G  & \gc \href{https://github.com/yaoching0/GaC}{\textcolor{gray}{[Link]}}\\
&
DeePEn~\cite{huang2024ensemble} &Token  & Aggregation,Vocabulary align.  &\cmark   &$\geq 2$  & Relative representation &  \xmark&OE-G/EM-G  & \href{https://github.com/OrangeInSouth/DeePEn}{\textcolor{gray}{[Link]}}\\
&
\gc EVA~\cite{xu2024bridging}   & \gc Token & \gc Averaging agg.,Vocabulary align.  & \gc \cmark  & \gc $\geq 2$    & \gc Relative representation  &  \gc \xmark & \gc OE-G/EM-G  & \gc \href{https://github.com/xydaytoy/EVA}{\textcolor{gray}{[Link]}}\\
&
UniTe~\cite{yao2024determine} &Token  &Averaging agg.,TOP-K union   & \cmark  &$\geq 2$  & Union dictionary &\cmark  & OE-G/EM-G & \textcolor{gray}{-}\\
&
\gc PackLLM~\cite{mavromatis2024pack} &\gc Token  &\gc Weighted avg. agg.,Perplexity   & \gc \cmark &\gc $\geq 2$  & \gc Greedy optimization &\gc \xmark  &\gc OE-G/EM-G  & \gc \href{https://github.com/cmavro/PackLLM}{\textcolor{gray}{[Link]}}\\
 &
 \cite{li2024purifying} & Token  & Weighted-averaging agg.   & \cmark  &  $=2$ &   Reduce negative issues  & \xmark   &  EM-G  &   \textcolor{gray}{-}\\
 &\gc
DeRa~\cite{liu2024decoding}  &\gc  Token  &\gc Weighted-averaging agg.   &\gc \cmark   &\gc$=2$  &\gc Decoding-time realignment  &\gc \xmark  &\gc OE-G/EM-G  &\gc \href{https://github.com/liutianlin0121/decoding-time-realignment}{\textcolor{gray}{[Link]}}\\

&
   MOD~\cite{shi2024decoding} &     Token &    Weighted-averaging agg.  &    \cmark  &    $\geq 2$ &    Multi-objective
decoding &    \xmark &    OE-G/EM-G &    \href{https://github.com/srzer/MOD}{\textcolor{gray}{[Link]}}\\

 & \gc
 Copilot~\cite{zou2025transformer}  & \gc Token  & \gc Weighted avg. agg.,Boosting   & \gc \xmark  & \gc  $= 2$   & \gc Token-level boosting,SFT  & \gc \xmark  & \gc OE-G/EM-G    & \gc \href{https://github.com/jiaruzouu/TransformerCopilot}{\textcolor{gray}{[Link]}}\\

  &  
 LLMBoost~\cite{chen2025llmboost}  &   Token  &   Weighted avg. agg.,Boosting   &   \xmark  &    $\geq 2$  &   Token-level boosting,SFT  &   \xmark  &   OE-G/EM-G    &    \textcolor{gray}{-}  \\

 &
 \gc  UltraFuser~\cite{ding2024mastering}   & \gc   Token    & \gc    Weighted avg. agg.,Gating 
        & \gc   \xmark   & \gc    $\geq 2$    & \gc    Gating mechanism,SFT    & \gc   \xmark   & \gc    OE-G/EM-G    & \gc     \textcolor{gray}{-}  
  \\

 &  
CDS~\cite{jin2024collaborative}  &   Token   &    Token-Level routing (selection)  &  \xmark   &   $=2$  &    Critical token classifier &  \xmark    &   EM-G     &    \textcolor{gray}{-}\\

  & \gc
  Co-Llm~\cite{shen2024learning}   & \gc   Token   & \gc    Token-Level routing (selection)   & \gc   \xmark    & \gc  $=2$   & \gc   Classifer,PGM   & \gc   \xmark    & \gc 
 
OE-G/EM-G 
   & \gc   \href{https://github.com/clinicalml/co-llm}{\textcolor{gray}{[Link]}}\\
 
 &   
CITER~\cite{zheng2025citer} &    Token &   Token-Level routing (selection)   &   \xmark  &   $= 2$ &   Reinforcement learning  &   \cmark  &    EM-G &    \href{https://github.com/aiming-lab/CITER}{\textcolor{gray}{[Link]}}\\

  & \gc
  ABE~\cite{wicks2025token}   & \gc   Token   & \gc   Token-Level routing (selection)     & \gc    \cmark   & \gc  $=2$   & \gc   Agreement-based ensembling   & \gc   \xmark    & \gc   OE-G/EM-G   & \gc   \href{https://github.com/mjpost/abe}{\textcolor{gray}{[Link]}}
\\
\cmidrule{2-10} 

\multirow{4}{*}{(b2)}
&    
Cool-Fusion~\cite{liu2025cool} &   Span  &    Generation-assessment-selection  &     \cmark &     $\geq 2$ &    Perplexity, common words &    \xmark  &    OE-G/EM-G  &    \textcolor{gray}{-}\\
  & \gc
  SweetSpan~\cite{xu2025hit}    & \gc    Span    & \gc     Generation-assessment-selection   & \gc   \cmark   & \gc  $  \geq 2$   & \gc   Perplexity, fixed-length   & \gc   \xmark    & \gc   OE-G/EM-G    & \gc   \textcolor{gray}{-}\\
&    
 SpecFuse~\cite{lv2024specfuse}  &    Span  &     Generation-assessment-selection   &      \cmark &      $\geq 2$ &     Perplexity, fixed-length &      \cmark &     OE-G/EM-G  &      \textcolor{gray}{-}
\\
  & \gc
 CoS~\cite{fu2025fast}    & \gc Span    & \gc Collaborative
decoding     & \gc  \cmark   & \gc  $\geq 2$   & \gc Speculation decoding   & \gc  \cmark   & \gc OE-G/EM-G    & \gc  
\href{https://github.com/Kamichanw/CoS/}{\textcolor{gray}{[Link]}}
\\



\cmidrule{2-10} 
\multirow{1}{*}{(b3)}
&    
 LE-MCTS~\cite{park2025ensembling} &     Process  &      Reasoning process selection  &     \xmark  &     $\geq 2$  &      Monte Carlo Tree 
Search &      \xmark  &     OE-G/EM-G &     \textcolor{gray}{-} \\


\bottomrule
\end{tabular}
}
\begin{tablenotes}
\small
\item[1] $\dag$:  It denotes to the number of models that can be used in the ensemble.
\item[2] $\S$: 
``Other Traits$^{\S}$'' denotes
unique and relatively important capabilities, modules or characteristics that are noteworthy.
\item[3] $^{*}$: OE-G denotes Open-Ended Generation; EM-G denotes Exact-Match Generation, characterized by objectively verifiable answers (e.g., mathematical solutions).
\end{tablenotes}
\end{threeparttable} 
}
\smallskip
\hfil
\captionsetup{type=table,skip=0pt}
  \caption{Summary of ensemble-during-inference methods.}
    \label{tab:Ensemble-during-inference}
\end{table*}

\subsection{Ensemble During Inference}
\label{Ensemble-during-inference}

We present a comprehensive review of \textit{token-level}, \textit{span-level}, and \textit{process-level} ensemble methods in the following subsections and offer a summary analysis in Table~\ref{tab:Ensemble-during-inference}.

\subsubsection{3.2.1 (b1) Token-Level Ensemble}
\label{Token-level Ensemble}

For these methods, at each decoding time-step, 1) \textit{aggregation} approach  produces the final token distribution for generation by (weighted) averaging  probability distributions from multiple LLMs, whereas 2) \textit{selection} 
approach opts to directly adopt the output token from a selected single model.

\paragraph{Aggregation-based normal methods.}
When attempting token-level aggregation, one encounters a dilemma---the \textit{vocabulary discrepancies} across different LLMs, which produce varying embedding lengths, impede the fusion and averaging of multiple probability distributions, creating a substantial obstacle during ensemble.
Thus, several methods focus on addressing the \textit{vocabulary alignment} issue.
\textbf{\textit{(i)}}
GaC~\cite{yu2024breaking} is the most straightforward method, which constructs a new   ``union dictionary'' by combining the vocabularies of multiple models to include all tokens from each dictionary, and subsequently projects the distribution information derived from each model onto this new merged dictionary for averaging aggregation;
\textbf{\textit{(ii)}}
Further, DeePEn~\cite{huang2024ensemble} and EVA~\cite{xu2024bridging} project the output distributions of multiple models at the current time step into a shared \textit{relative}/\textit{pivot space}, followed by either an averaging aggregation~\cite{huang2024ensemble} or a weighted aggregation~\cite{huang2024ensemble,xu2024bridging} to derive a final  distribution.
\footnote{Beyond the aforementioned research focusing on vocabulary alignment, all other token-level aggregation-based ensemble methods  either employ existing alignment techniques or assume that the LLMs under consideration share a common vocabulary.}
Furthermore, there are additional advanced aggregation-based token-level ensemble methods.
Specifically, 
\textbf{\textit{(i)}}
Yao et al.~\shortcite{yao2024determine} propose UniTe, which focuses solely on the TOP-K portion of each model’s output distribution 
for performing averaging aggregation,  
reducing computational overhead;
\textbf{\textit{(ii)}}
Mavromatis et al.~\shortcite{mavromatis2024pack} present a \textit{weighted} averaging approach, where the weights were determined by the \textit{perplexity} 
of each model at each time step, indicating whether  the current input aligns with the respective model expertise.

\paragraph{Aggregation-based specific-goal methods.}
Unlike the aggregation-based normal methods introduced above, which are primarily intended to improve task performance in general scenarios, there are other methods that have their own specific goals.
These methods aim to mitigate the prominent weaknesses of a single large LLM~\cite{li2024purifying} or achieve a balanced effects (including decoding-time realignment ~\cite{liu2024decoding} and multi-objective decoding~\cite{shi2024decoding}) by \textit{weighted averaging} the token-level outputs from multiple LLMs with different characteristics.
Among them, Li et al. ~\shortcite{li2024purifying} reduce negative issues during the deployment of LLMs, such as copyright infringement and data poisoning, by integrating the token-level outputs of a benign small model with a large model using weighted averaging aggregation. 

\paragraph{Aggregation-based fine-tuning methods.}
Additionally, there are methods such as Copilot~\cite{zou2025transformer}, LLMBoost~\cite{chen2025llmboost} and UltraFuser~\cite{ding2024mastering}, which consider task-downstream supervised fine tuning. 
These methods  use labeled data to fine-tune multiple pre-trained LLMs based on the Ensemble Learning paradigm~\cite{zhou2021ensemble}, i.e., boosting~\cite{zou2025transformer,chen2025llmboost} and MOE (Mixture-of-Experts)~\cite{ding2024mastering}. 
Finally, they perform token-level weighted averaging ensemble during the inference phase.

\paragraph{Selection-based methods.}
These methods directly choose a token from a specific model's output. 
\textbf{\textit{(i)}}
Jin et al.~\cite{jin2024collaborative} propose CDS, which trains a binary classifier based on the preceding context. This classifier determines whether to invoke an unaligned model (to leverage its factual knowledge) or an aligned model (to utilize its instruction-following and safety capabilities) at each decoding step;
\textbf{\textit{(ii)}} Similar to CDS, Co-Llm~\cite{shen2024learning} employs a routing classifier to switch between a general-purpose model and a specialized model for optimal output;
\textbf{\textit{(iii)}} Similarly, Zheng et al.~\shortcite{zheng2025citer} introduce CITER, which uses reinforcement learning to train a policy function; 
\textbf{\textit{(iv)}} 
Wicks et al.~\cite{wicks2025token} use an approach they termed ``agreement-based ensembling'' to select the intersection of various token information.

\subsubsection{3.2.2 (b2) Span-Level Ensemble}
\label{Span-level Ensemble}

\textbf{\textit{(i)}} Existing span-level ensemble methods mostly adhere to a ``generation-assessment-selection'' pipeline. 
These methods first employ multiple large models to generate word-containing fragments, then utilize the concept of \textit{perplexity}, to make each model score all generated model responses, ultimately selecting the fragment with the highest cumulative score.
The key distinction lies in that Gool-Fusion~\cite{liu2025cool}
have each source LLM individually generate text segments until
the word boundary of each segment is common to all LLMs, whereas SweetSpan~\cite{xu2025hit} and SpaceFuse~\cite{lv2024specfuse} adopt a more straightforward approach by setting the span as a fragment with a fixed word count; 
\textbf{\textit{(ii)}}
Further, Fu et al.~\shortcite{fu2025fast} incorporate \textit{speculative decoding}, a well-known acceleration technique in the LLM field, into the  ensemble process to achieve faster inference.

\subsubsection{3.2.3 (b3) Process-Level Ensemble}
\label{Process-level Ensemble}

Confronted with complex step-by-step reasoning tasks, Park et al.~\shortcite{park2025ensembling} use a trained Monte Carlo Tree Search strategy at each reasoning step to select the output with the highest reward value from multiple model reasoning outputs, thereby identifying the most accurate reasoning chain.

\setlength{\cmidrulewidth}{0.01em}

\begin{table*}[t]
\hspace*{0.1cm}  
\scalebox{0.63}{
\fontsize{9pt}{9pt}

\begin{threeparttable}  
\scalebox{0.93}{
\begin{tabular}{llcc c c c c c }
\toprule
& \textbf{Methods}    &  \textbf{Un-/Supervised} &\textbf{Main Strategies}   & \textbf{\# Models}  & \textbf{Other Traits$^{\S}$}  & \textbf{Efficiency-aware}  & \textbf{Tasks}$^{*}$ & \textbf{Code} \\

\midrule
\multirow{7}{*}{(c1)}
&\gc  Agent-Forest~\cite{li2024more}  & \gc  Unsupervised & \gc  Similarity-based selection & \gc  $ >  2$& \gc   Scaling property & \gc  \xmark &  \gc  OE-G/EM-G & \gc  \href{https://github.com/MoreAgentsIsAllYouNeed/AgentForest}{\textcolor{gray}{[Link]}}\\

& Smoothie~\cite{guhasmoothie} & Unsupervised& Similarity-based selection  & $ >  2$  & Graphical model &\xmark  &OE-G/EM-G  & \href{https://github.com/HazyResearch/smoothie}{\textcolor{gray}{[Link]}}\\

& \gc LLM-PeerReview~\cite{chen2025scoring}  & \gc Unsupervised & \gc PeerReview framework  & \gc $ >  2$  & \gc Probabilistic Graphical Model & \gc \xmark & \gc OE-G/EM-G & \gc\href{https://github.com/zeyuji/LLM-PeerReview}{\textcolor{gray}{[Link]}}
\\\  
&  MoRE~\cite{si2023getting} &   Supervised &    
Supervised Sim.-based sel. &   $ >  2$  &   Feature engineering &   \xmark&    EM-G &   \href{https://github.com/NoviScl/MoRE}{\textcolor{gray}{[Link]}}
\\

& \gc LLM-Blender~\cite{jiang2023llm}  & \gc  Supervised & \gc  Selection-then-regeneration  & \gc  $ >  2$  & \gc  Pairwise ranking& \gc  \xmark & \gc  OE-G/EM-G & \gc \href{https://github.com/yuchenlin/LLM-Blender}{\textcolor{gray}{[Link]}}\\\  

&  LLM-TOPLA~\cite{tekin2024llm}  &   Supervised &   Selection-then-regeneration  &   $ >  2$  &     Maximising diversity &   \xmark &   OE-G/EM-G &   \href{https://github.com/git-disl/llm-topla}{\textcolor{gray}{[Link]}}\\

& \gc URG~\cite{lv2024urg}  & \gc  Supervised & \gc  Selection-then-regeneration & \gc  $ >  2$ & \gc  Unified processes & \gc \cmark & \gc  OE-G/EM-G & \gc  -\\
\cmidrule{2-9} 
\multirow{10}{*}{(c2)}  &   EcoAssistant~\cite{zhang2023ecoassistant} 	&   Unsupervised &   User judgment &   $\geq 2$ &   Prompt engineering &   \cmark &   OE-G (code generation) &   \href{https://github.com/JieyuZ2/EcoAssistant}{\textcolor{gray}{[Link]}}\\
& \gc   \cite{yuelarge}  & \gc  Unsupervised  & \gc  Answer consistency & \gc  $\geq 2$$^{\dag}$ & \gc Sampling and checking & \gc \cmark  & \gc EM-G & \gc  \href{https://github.com/MurongYue/LLM_MoT_cascade}{\textcolor{gray}{[Link]}}\\
&    Model Cascading~\cite{varshney2022model}	&   Unsupervised &   Class uncertainty &    $\geq 2$  &   Cascade-pioneer &  \cmark &   EM-G &   \textcolor{gray}{-}\\
& \gc    neural caching~\cite{ramirez2023cache}	& \gc  Unsupervised & \gc  Class unc./Ans. cons. & \gc  $= 2$  & \gc  Distillation  & \gc  \cmark & \gc EM-G  & \gc  \href{https://github.com/guillemram97/neural-caching}{\textcolor{gray}{[Link]}} \\

&    Cascade Routing~\cite{dekoninck2024unified}	&   Unsupervised &   Class uncertainty  &   $>2$  &  Routing-equipped &  \cmark  &  EM-G  &  \href{https://github.com/eth-sri/cascade-routing}{\textcolor{gray}{[Link]}} \\

& \gc   \cite{jitkrittum2024does}	& \gc  Supervised & \gc  (Upgraded) Class unc. & \gc  $\geq2$$^{\dag}$ & \gc  Post-hoc deferral & \gc  \cmark & \gc  EM-G & \gc  \textcolor{gray}{-} \\

&    FrugalGPT~\cite{chenfrugalgpt} 	&   Supervised &    Score function  &  $\geq2$&   Prompt engineering, etc.&   \cmark &   EM-G  &  \textcolor{gray}{-}\\

& \gc   \cite{gupta2024language} 	& \gc  Supervised & \gc  Score function 
& \gc  $=2$ & \gc  Quantile features & \gc  \cmark & \gc  OE-G/EM-G & \gc  \textcolor{gray}{-}\\

&    AutoMix~\cite{aggarwal2023automix}	&   Supervised &   MDP &  $\geq2$ &  Routing-equipped, etc.&  \cmark &  EM-G &  \textcolor{gray}{-}\\

& \gc   DER~\cite{hu2024dynamic}  & \gc  Supervised & \gc  MDP,Ans. Cons. & \gc $\geq2$ & \gc Routing-equipped& \gc \cmark & \gc EM-G& \gc \textcolor{gray}{-} \\
\bottomrule
\end{tabular}
}
\begin{tablenotes}
\small
\item[1] $\S$: 
``Other Traits$^{\S}$'' denotes 
unique and relatively important capabilities, modules or characteristics that are noteworthy.
\item[2] $^{*}$: OE-G denotes Open-Ended Generation; EM-G denotes Exact-Match Generation, characterized by objectively verifiable answers (e.g., mathematical solutions).
\item[3] $\dag$: ``$\geq2$$^{\dag}$'' denotes  that the study considers a 2-models cascade scenario in its methodology introduction/derivation, but it can be easily adapted to K-models cascade scenarios.
\end{tablenotes}
\end{threeparttable} 
}
\smallskip
\hfil
\captionsetup{type=table,skip=0pt}
  \caption{Summary of ensemble-after-inference methods.}
    \label{tab:Ensemble-after-inference}
\end{table*}

\subsection{Ensemble After Inference}
\label{Ensemble-after-inference}

We review the \textit{non-cascade} and \textit{cascade} methods in the following, and provide a summary analysis in Table~\ref{tab:Ensemble-after-inference}.

\subsubsection{3.3.1 (c1) Non-Cascade}

As shown in Figure~\ref{Fig: Taxonomy} and Table~\ref{tab:Ensemble-after-inference}, existing non-cascade methods fall into two categories: 1) \textit{selection-based}, which is dedicated to selecting a single response from multiple candidates,
and 2) \textit{selection-then-regeneration}, which involves initially selecting a subset of candidate responses and subsequently feeding this refined subset into a generative model for regeneration to obtain the final output.

\paragraph{Selection-based methods.}
First, as \textit{unsupervised} methods, Agent-Forest~\cite{li2024more}  and Smoothie~\cite{guhasmoothie}  essentially leverage the similarity between model responses for selection and adhere to a majority voting (MV) principle:
\textit{selecting the response that exhibits the highest degree of similarity to the others as the final answer}. 
Specifically, \textbf{\textit{(i)}}
Li et al.~\shortcite{li2024more} obtain multiple responses by repeatedly querying a single model (equivalent to utilizing outputs from multiple homogeneous models) and employ the MV ensemble, discovering a scaling property associated with ensembling more responses; they use BLEU scores and occurrence frequencies to quantify the similarities between responses for open-ended and exact-match generation tasks, respectively;
\textbf{\textit{(ii)}}
Guha et al. ~\shortcite{guhasmoothie} input multiple $<$query, response$>$ pairs into  SentenceBERT, leveraging the Euclidean distance between the resulting low-dimensional encoded vectors as a measure of similarity.
Chen et al.~\shortcite{chen2025scoring} propose LLM-PeerReview, an academic peer-review-inspired framework that leverages LLM-as-a-Judge to let LLMs score the quality of responses and select the response with the highest final aggregated score as the output.

Further, as a \textit{supervised} method, MoRE~\cite{si2023getting} utilizes some training data to train a random forest classifier for selecting among multiple responses, using the response similarity among LLMs as one of the constructed features.

\paragraph{Selection-then-regeneration methods.}
These methods stem from the pioneering work LLM-Blender, introduced by Jiang et al.~\shortcite{jiang2023llm}.
\textbf{\textit{(i)}}
LLM-Blender uses some training data to train the ``PairRanker'' selection module in the first phase, serving to select a subset from multiple responses,  and to train the ``GenFuser'' generator in the subsequent phase for synthesizing the final response;
\textbf{\textit{(ii)}}
Building upon LLM-Blender, the improvement made by LLM-TOPLA~\cite{tekin2024llm} mainly lies in optimizing the selection process in the first step by ``maximizing the diversity among multiple response candidates'';  
\textbf{\textit{(iii)}}
In addition, URG~\cite{lv2024urg} proposes an end-to-end framework that integrates selection and regeneration.

\subsubsection{3.3.2 (c2) Cascade}
Methodologically, all cascade methods revolve around the \textit{deferral rule}/\textit{decision maker}: determining whether to adopt the output of the current model at hand or to invoke a subsequent, more powerful model.


\paragraph{Unsupervised methods.}
As shown in Table~\ref{tab:Ensemble-after-inference}, these methods can be further categorized into three types according to the pivotal ``main strategy'': \textit{user judgment}~\cite{zhang2023ecoassistant}, \textit{answer consistency}~\cite{yuelarge,ramirez2023cache}, and \textit{class uncertainty}~\cite{varshney2022model,ramirez2023cache,dekoninck2024unified}.

As one most straightforward method, Zhang et al. \shortcite{zhang2023ecoassistant}  propose a cascade method, EcoAssistant, that utilizes \textit{user judgment} rather than \textit{machine algorithms} to determine 
whether the cascading inference should be terminated.
Further, Yue et al.~\shortcite{yuelarge} introduce a method based on \textit{answer consistency}—more precisely, on the observation of whether the inference results generated from multiple identical/diverse prompts exhibit sufficient uniformity—to solve the cascade problem for in-context-learning reasoning tasks.


Last but certainly not least, a critically important category of approaches is grounded in \textit{class uncertainty}------an approach that also manifests in the following ``supervised methods''. 
Class uncertainty, sometimes also referred to as \textit{confidence},
can be implemented through various variants, including Maximum Softmax Probability~\cite{varshney2022model}, Distance To Uniform Distribution~\cite{varshney2022model}, Margin Sampling~\cite{ramirez2023cache}, and Prediction Entropy~\cite{ramirez2023cache}. 
However,  at its core, it evaluates whether the probability of the dominant class inferred by the current model exceeds a sufficient threshold.
When this probability is sufficiently high---signifying adequate confidence---the cascading process is concluded; otherwise, it proceeds.

\paragraph{Supervised methods.}
Unlike the aforementioned unsupervised cascade methods, supervised cascade methods use some training data to train cascade-related modules, including \textit{post-hoc deferral based on upgraded class-uncertainty strategy}~\cite{jitkrittum2024does}, \textit{scoring functions}~\cite{chenfrugalgpt,gupta2024language}, and \textit{Markov Decision Processes (MDPs)} ~\cite{aggarwal2023automix,hu2024dynamic}.
Specifically, \textbf{\textit{(i)}} Jitkrittum et al.~\shortcite{jitkrittum2024does} use the training data to learn a cascade method called \textit{post-hoc deferral}; and the core idea is that, in contrast to aforementioned \textit{class-uncertainty}
cascade methods
which solely focus on the class uncertainty of the current model’s output, it also takes into account the estimated uncertainty of the next, stronger model’s output;
\textbf{\textit{(ii)}} Further, Chen et al.~\shortcite{chenfrugalgpt} and Gupta et al.~\shortcite{gupta2024language} use the training data to learn a \textit{scoring function} that produces a confidence score regarding the current model’s output, enabling cascading judgments;
\textbf{\textit{(iii)}} 
Both Aggarwal et al.~\shortcite{aggarwal2023automix} and Hu et al.~\shortcite{hu2024dynamic} adopt theidea of ``introducing routing in the cascading process'' and train an MDP for decision-making.

\setlength{\cmidrulewidth}{0.01em}  

\begin{table*}[t!]
\hspace*{-0.1cm}  
\scalebox{0.89}{
\fontsize{9pt}{9pt}

\begin{threeparttable}  
\scalebox{0.94}{
\begin{tabular}{llcc    c}
\toprule
 & \textbf{Approaches}    &  \textbf{Ensemble Strategies} &\textbf{Ensemble Granularities}     & \textbf{Ensemble Goals}   \\

\midrule

\multirow{2}{*}{(a) Ensemble before inference}

 & (a1) Discrete utility methods &Selection  & Response-level     \({\tiny\clubsuit}\)    & Performance (and cost) \\

 & (a2) Continuous utility methods &Selection  & Response-level \({\tiny\clubsuit}\)  &Performance and cost   \\

\cmidrule{1-5} 
\multirow{3}{*}{(b) Ensemble during inference}
 & (b1) Token-level ensemble & 
 Aggregation, Selection  & Token-level     \({\tiny\clubsuit}\)  \({\tiny\clubsuit}\)  \({\tiny\clubsuit}\)   &Performance$^{\S}$  \\
 & (b2) Span-level ensemble &Selection  & Span-level    \({\tiny\clubsuit}\)  \({\tiny\clubsuit}\)  &Performance \\

 & (b3) Process-level ensemble &Selection  & Process-level   \({\tiny\clubsuit}\)  \({\tiny\clubsuit}\)    &Performance   \\

\cmidrule{1-5} 

\multirow{2}{*}{(c) Ensemble after inference}
 &(c1) Non cascade &  Selection, Regeneration  & Response-level   \({\tiny\clubsuit}\) &Performance   \\
 &(c2) Cascade &Selection  & Response-level  \({\tiny\clubsuit}\)  &Performance and cost   \\

\bottomrule
\end{tabular}
}
\begin{tablenotes}
\small
\item[1] \({\tiny\clubsuit}\): It represents the level of granularity, and a higher quantity indicates finer ensemble granularity.
\item[2] $^{\S}$: As a specific example, Li et al. ~\shortcite{li2024purifying} primarily aim to minimize the negative issues of large models during deployment.
\end{tablenotes}
\end{threeparttable} 
}
\smallskip
\hfil
\captionsetup{type=table,skip=5pt}
  \caption{Summary analysis of the key attributes of LLM Ensemble approaches.}
    \label{summary-final}
\end{table*}

\section{Benchmarks and Applications}

In this section, we briefly discuss popular benchmarks and applications in LLM Ensemble.
First, there are two types of benchmarks specifically tailored for LLM Ensemble evaluation. 
1) Benchmark \textsc{MixInstruct} proposed by Jiang et al. \shortcite{jiang2023llm}, which serves to assess the \textit{performance} of ensemble-after-inference methods;
2) In contrast, most later benchmarks target ensemble-before-inference, i.e., LLM routing: 
RouterEval~\cite{huang2025routereval} is mainly performance-oriented, 
RouterBench~\cite{hu2024routerbench}, FusionFactory~\cite{feng2025fusionfactory}, RouterArena~\cite{lu2025routerarena} and LLMRouterBench~\cite{li2026llmrouterbench} further evaluate performance–cost. 
In addition, in terms of applications, beyond the methods outlined in Section~\ref{Methodology}, the concept of LLM Ensemble has found applications in a variety of more specialized tasks and domains.  For instance, 
Lee et al. ~\shortcite{lee2023ensemble} leverage the ROUGE-L metric to assess the similarity of generated text and employed a similarity-based selection strategy for ensemble to produce Instruction-Tuning data; 
additionally, several studies focus on tabular data imputation~\cite{he2025llm}, win rate evaluation~\cite{gao2024bayesian}, SQL generation~\cite{gundabathula2024promptmind},  tool routing for RAG systems~\cite{mu2024adaptive}, and so on.

\section{Discussion}

\subsection{Summarization}

Here we provide a summary analysis of all LLM Ensemble approaches, as illustrated in Table~\ref{summary-final}, considering the most critical methodological attributes:
\textit{ensemble strategy}, \textit{granularity} and \textit{ensemble goal}.
From the perspective of ensemble strategy, \textit{aggregation} methods (involving the average or weighted ensemble of all model outputs) are more sophisticated compared to \textit{selection}-based methods, which select a single output (equivalent to a \textit{hard voting} process). 
However, \textit{regeneration}-based methods are burdened by the additional need for large model-specific training data preparation and model training.
From the perspective of ensemble granularity, it is apparent that \textit{response}-level ensemble methods are relatively coarse-grained; while methods with finer granularity, particularly token-level ensemble methods, can
more effectively harness the distribution information from each model during the decoding phase.
Further, from the perspective of ensemble goals, (b) ensemble-during-inference methods and (c1) non-cascaded-based methods, unconstrained by cost considerations, can employ more flexible ensemble strategies beyond \textit{selection} and utilize finer-grained ensemble approaches, ultimately yielding greater performance improvement potential.

\subsection{Existing Limitations and Future Directions}

\begin{myboxi}[Summary on Future Directions]
\begin{itemize}
    \item 1) Label-efficient un-/weakly-supervised model profiling for ensemble-before-inference methods.
    \item  2)
    Efficient adaptation under evolving model distributions for ensemble-before-inference.
    \item  3)
    Principled span-level ensemble-during-inference approach. 
    \item  4)
    Sophisticated unsupervised cascade ensemble-after-inference approach for open-ended generation.
\end{itemize}
\end{myboxi}

\paragraph{Direction 1.}
To estimate model competence across domains, existing ensemble-before-inference methods typically rely on supervised profiling data that records the correctness of model outputs, requiring non-trivial labeling effort and domain-specific annotation resources. Intuitively, a model’s internal state may already encode informative signals about its domain-specific expertise and reliability, which could be exploited for label-efficient profiling. To this end, a promising direction is to develop un-/weakly-supervised profiling strategies that infer model strengths from intrinsic signals—e.g., prediction uncertainty (entropy), self-consistency, or internal representations—thereby enabling low-cost model profiling without extensive labeled data.

\paragraph{Direction 2.} 
Existing ensemble-before-inference methods are typically evaluated in closed and static settings with a fixed model pool. However, real-world deployments are dynamic: the set of available models and their capabilities continuously evolve, leading to shifts in the candidate model distribution. This calls for routing strategies that can \emph{rapidly and data-efficiently} adapt to evolving model distributions, enabling low-cost few-shot recalibration when models are added, removed, or updated.

\paragraph{Direction 3.} 
Span-level ensemble-during-inference methods already offer sufficient granularity and show higher performance potential compared to response-level methods~\cite{xu2025hit}. 
However, current segment segmentation techniques are still overly simplistic (e.g., rigidly defining segment lengths as  a 4-word sequence).
Shifting to a more principled segment segmentation approach could provide richer, more valuable information for the subsequent ensemble process.

\paragraph{Direction 4.} 
Existing cascade methods, compared to ensemble-before-inference methods, can also address the cost-considered ensemble problem, with the added benefit of utilizing model responses during the cascade process to select the most appropriate output.
However, most existing cascade methods are unable to handle open-ended generation tasks or depend on supervised learning with labeled data, thus losing  generalizability.
Therefore, developing a general unsupervised cascade approach would be of great significance.

\section{Conclusion}

LLM Ensemble stands as a direct manifestation of ensemble learning in the era of large language models. 
The accessibility, diversity, and out-of-the-box usability of large language models have made the spirit of ensemble learning shine even brighter, fueled by the rapidly developing LLM Ensemble studies.
This paper provides a comprehensive taxonomy and a review of the methods in the field of LLM Ensemble, introduces relevant applications and benchmarks, conducts a summative analysis of these methods, and proposes several potential research directions.
We hope that this review will offer valuable insights to researchers and inspire further advancements in LLM Ensemble and related research areas.



\bibliographystyle{named}
\bibliography{output}

\end{document}